\pdfoutput=1
\documentclass[11pt]{article}

\usepackage[final]{acl}

\usepackage{times}
\usepackage{latexsym}
\usepackage[T1]{fontenc}

\usepackage[utf8]{inputenc}

\usepackage{microtype}

\usepackage{inconsolata}

\usepackage{graphicx}

\usepackage[utf8]{inputenc} 
\usepackage[T1]{fontenc}    
\usepackage{hyperref}       
\usepackage{url}            
\usepackage{booktabs}       
\usepackage{amsfonts}       
\usepackage{nicefrac}       
\usepackage{microtype}      
\usepackage{xcolor}         
\usepackage{float}
\usepackage[utf8]{inputenc}
\usepackage{newtxtext,newtxmath} 
\usepackage{graphicx} 
\usepackage{enumitem}
\usepackage{multirow}
\usepackage{xspace}
\usepackage[most]{tcolorbox}
\usepackage{xcolor}
\usepackage{pifont}
\usepackage{cleveref}
\usepackage{array}
\usepackage[normalem]{ulem}
\newcommand{\resolved}[1]{}
\usepackage{etoolbox}

\graphicspath{ {./figures/} }

\newcommand{\sftmodel}{\text{StresSLM}\xspace}
\newcommand{\stressds}{\textit{StressTest}\xspace}
\newcommand{\stresspresso}{\textit{StressPresso}\xspace}
\newcommand{\traindata}{\text{Stress-17k}\xspace}
\newcommand{\whistress}{\text{WhiStress}\xspace}
\newcommand{\ssd}{\text{SSD}\xspace}
\newcommand{\ssr}{\text{SSR}\xspace}
\newcommand{\cmark}{\ding{51}}  
\newcommand{\xmark}{\ding{55}}  

\newtcolorbox{styleprompt}[1][]{
  enhanced,
  breakable,
  colback=blue!5!white,
  colframe=blue!50!black,
  coltitle=blue!30!black,
  colbacktitle=blue!10!white,
  title=LLM Prompt,
  fonttitle=\bfseries,
  boxrule=0.8pt,
  arc=6pt,                       
  top=1mm, bottom=1mm,
  left=2mm, right=2mm,
  #1
}

\title{\stressds: Can YOUR Speech LM Handle the Stress?}

\author{Iddo Yosha ~~~~~ Gallil Maimon ~~~~~ Yossi Adi\\ \\ School of Computer Science and Engineering \\ The Hebrew University of Jerusalem\\\\\texttt{iddo.yosha@mail.huji.ac.il}}

\begin{document}
\maketitle

\begin{abstract}
Sentence stress refers to emphasis on words within a spoken utterance to highlight or contrast an idea. It is often used to imply an underlying intention not explicitly stated. Recent speech-aware language models (SLMs) have enabled direct audio processing, allowing models to access the full richness of speech to perform audio reasoning tasks such as spoken question answering. Despite the crucial role of sentence stress in shaping meaning and intent, it remains largely overlooked in evaluation and development of SLMs. We address this gap by introducing \stressds, a benchmark designed to evaluate models' ability to distinguish between meanings of speech based on the stress pattern. We evaluate leading SLMs, and find that despite their overall capabilities, they perform poorly on such tasks. Hence, we propose a novel data generation pipeline, and create \traindata, a training set that simulates change of meaning implied by stress variation. Results suggest, that our finetuned model, \sftmodel, generalizes well to real recordings and notably outperforms existing SLMs on sentence stress reasoning and detection. Models, code, data, samples - \href{https://pages.cs.huji.ac.il/adiyoss-lab/stresstest}{https://pages.cs.huji.ac.il/adiyoss-lab/stresstest}.
\end{abstract}

\section{Introduction}
\label{sec:intro}
Large language models (LLMs) have revolutionized language processing, enabling new forms of human-computer interaction \citep{minaee2025largelanguagemodelssurvey, grattafiori2024llama3herdmodels}. Further research explored integration of other modalities into LLMs. Notably, incorporation of speech and audio into LLMs has gained traction, equipping models with the ability to listen, speak, and reason about audio \citep{arora2025landscapespokenlanguagemodels, ghosh2025audioflamingo2audiolanguage, maimon2025scaling}.

An elementary approach of integrating speech into LLMs follows a cascade paradigm~\citep{ji2024wavchat}, where audio is transcribed by an automatic speech recognition (ASR) system, and the resulting text is processed by an LM. While somewhat effective, this approach falls short of capturing the full expressive range of spoken language. Speech carries rich paralinguistic cues such as emotion, speaker identity, and prosodic characteristics that are often lost in transcription~\citep{WILSON20061559, AcousticCorrelates}.
One key aspect of prosody is sentence stress, which refers to the emphasis placed on particular words or phrases within a sentence to highlight an idea or to contrast another \citep{Bolinger}. For example, the sentence \textit{``I didn’t say she stole the money''} can express dramatically different meanings, based on the stressed words, for the same written-form.

Recent research on SLMs aim to address these limitations by enabling models to process audio directly, bypassing transcription and allowing them to access acoustic information~\citep{arora2025landscapespokenlanguagemodels}. Leading models in this space have demonstrated impressive performance across a wide range of speech tasks~\citep{chu2024qwen2audiotechnicalreport, tang2024salmonngenerichearingabilities}. Nevertheless, sentence stress has received limited attention in the evaluation and development of SLMs, despite its vital role in expressing the speaker’s intent and meaning. We argue that interpreting sentence stress requires the listener to reason about the intended meaning based on stress placement, which can often be inferred even without explicit context.

\begin{figure*}[t!]
    \centering
    \includegraphics[width=1\textwidth]{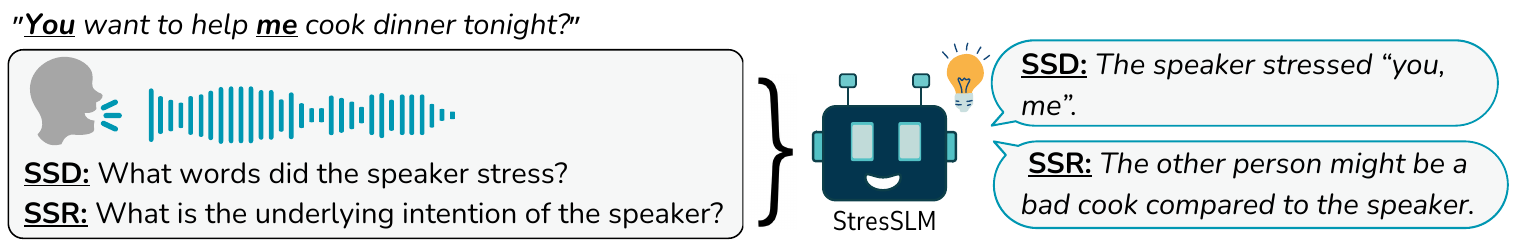}
    \caption{\stressds provides samples that can be understood differently based on stress. We consider sentence stress detection (\ssd) and sentence stress reasoning (\ssr). \sftmodel detects stress and reasons about the meaning. 
    \label{fig:bench_example}}
\end{figure*}

In this work, we address this gap by introducing \stressds, a comprehensive benchmark designed to evaluate a model's ability to distinguish spoken sentence meanings based on different stress patterns. A visual example can be seen in \Cref{fig:bench_example}. We then evaluate leading SLMs on our benchmark to quantify their capacity for stress based reasoning. Additionally, we introduce a novel synthetic dataset generation pipeline, which produces the \traindata dataset. We empirically show that finetuning a model using \traindata leads to enhanced ability to detect and model sentence stress on real-world recordings. Through extensive empirical evaluation and ablation studies, we demonstrate that our model, \sftmodel, significantly outperforms existing models in both stress detection and reasoning, with minimal performance drop on original tasks. 

\paragraph{Our contributions:} (i) We propose \stressds, a novel benchmark for evaluating sentence stress understanding in SLMs, and further extend it with existing stress-labeled data; (ii) We analyze performance of several leading SLMs on the benchmark, and show that they fail to detect stress and reason about its meaning; (iii) We propose a data synthesis and filtering pipeline and demonstrate its effectiveness for training an SLM for stress understanding.

\section{Background}
\label{background}
\paragraph{Theoretical views of sentence stress.} Theoretical accounts of sentence stress fall into two perspectives \citep{Ladd_2008}: (i) The phonological view that treats normal stress as a default prosodic pattern governed by syntax, formalized in the Nuclear Stress Rule \citep{Chomsky1968TheSP}. This approach is not tied to meaning; (ii) The semantic view, which considers stress as context-dependent, reflecting the speaker’s intent to mark focus, contrast, new information, or emphasis \citep{Bolinger}. In this work, we adopt the latter, asking whether SLMs can infer speaker-intended meaning from stress.

Within the semantic view, sentence stress can be grouped into four overlapping categories: (i) \emph{Contrastive Stress}, marking opposition \citep{ContrastiveAccentAndContrastiveStress, ContrastiveStatements, MeaningAndContrastiveStress}; (ii) \emph{Emphatic Stress}, amplifying or diminishing intensity, often with emotional or scalar meaning \citep{ALinguisticAnalysisOfSentenceStress}; (iii) \emph{New Information Stress}, signaling the introduction of novel or unexpected content \citep{stressAndInformation,ALinguisticAnalysisOfSentenceStress}; and (iv) \emph{Focus Stress}, highlighting the most relevant element for discourse \citep{ALinguisticAnalysisOfSentenceStress, Ladd_2008}. Examples can be see in \Cref{tab:stress_types}.

\begin{table*}[t!]
\centering
\small
\begin{tabular}
{@{}p{2.3cm}p{4.5cm}p{8.0cm}@{}}
\toprule
\textbf{Stress Type} & \textbf{Description} & \textbf{Stressed speech (intention)} \\
\midrule
\multirow{2}{*}{\textbf{Contrastive}} & Demonstrates contrast with another option. & 
\textit{``\underline{I} didn't take your book.''} (vs. someone else) \newline
\textit{``I didn't take your \underline{book}.''} (vs. something else) \\
\addlinespace
\multirow{2}{*}{\textbf{Emphatic}} & Amplifies or diminishes the intensity of a concept. &
\textit{``They \underline{loved} how you treated her dog.''} (You really exceeded their expectations) \\
\addlinespace
\multirow{2}{*}{\textbf{New-Information}} & Marks a surprising or novel content in the discourse. &
\textit{``He's actually moving to \underline{New York}.''} (surprising since its far from his current home) \\
\addlinespace
\multirow{2}{*}{\textbf{Focus}} &  General-purpose mechanism for highlighting key elements. &
\textit{``I enjoy the taste of espresso \underline{at sunrise}.''} (It's about that particular time) \\
\bottomrule
\end{tabular}
\caption{Examples of sentence stress categories from \stressds.}
\label{tab:stress_types}
\end{table*}

\paragraph{Realization of sentence stress in speech.} A complementary line of research has focused on how stress is realized in a speech signal. These studies investigate the acoustic aspects that contribute to the production and perception of stress. The most widely agreed-upon prosodic features associated with stress are pitch (f$0$), loudness, duration and vowel quality~\citep{AcousticCorrelates, ProsodicStressRevisited}. 

\section{Benchmarking stress understanding}
\label{sec:evaluation_benchmark}

In benchmarking stress understanding in SLMs, we focus on two tasks: (i) \emph{Sentence Stress Reasoning} (\ssr), evaluates whether an SLM can infer the speaker’s intended meaning from the speech alone. This task goes beyond recognizing what was said, it requires understanding how it was said, focusing on how stress influences the interpretation of an utterance; (ii) \emph{Sentence Stress Detection} (\ssd), evaluates the model’s ability to identify stressed word(s), given the ground truth transcription. This simpler task provides a controlled assessment of the model’s sensitivity to acoustic prominence. Note that, in contrast to \ssr which is a novel task, \ssd is established in literature. We include it as a complementary task to \ssr.

\subsection{Datasets}\label{sec:eval_ds}
\paragraph{\stressds} is a dataset of texts recorded in two (or more) stress patterns, implying different meanings. It is recorded by a single professional actor. Exact details about the number of recordings per text, and recording setup can be found in Appendix~\ref{appendix:benchmark}.

As not all texts can naturally accommodate multiple stress patterns, text sentences have to be collected specifically for this purpose. Therefore, text samples were curated through a rigorous manual process. Annotators who are fluent in English were instructed to suggest sentences that could have at least two different meanings based on a stressed word or words. We provide annotators with several examples from literature for inspiration, and they could use semi-automatic tools for proposing new examples (e.g. LLMs), but had to manually write and verify the samples. We also request they do not use exact templates of the examples, even if they use similar stress types. These samples were verified for agreement with at least one further annotator. In cases where the stressed words were agreed upon, but the meaning's text description needed refinement this was done by a third annotator. Finally, annotators were requested to mark samples with corrupted audio for removal. A breakdown of the various stress patterns in \stressds, along with their frequencies, is shown in \Cref{fig:stress_ven} in Appendix \ref{appendix:benchmark}.

\paragraph{\stresspresso.} We further extend our evaluation by annotating a subset of Expresso~\citep{nguyen2023expressobenchmarkanalysisdiscrete} — a real-speech stress-labeled dataset. We denote this subset as \stresspresso. As Expresso is an expressive speech dataset, we consider neutral-emotion samples only, to decouple emotion from stress understanding. A major distinction from \stressds lies in the annotation process. 
\stresspresso was annotated post-hoc, with annotators inferring underlying intentions from existing audio, whereas \stressds was annotated with an intended meaning and then recorded accordingly. The rest of the annotation procedure was the same. Another difference is that \stresspresso has four speakers, two male and two female. Further data statistics can be found in Appendix~\ref{appendix:benchmark}.

Importantly, \stresspresso serves as a complementary evaluation set to the primary benchmark. This is due to inherent limitations of the post-hoc annotation process. This limitation motivated the design of \stressds, where recordings were explicitly produced to match intended meanings, ensuring alignment between speech and annotation. At the same time, \stresspresso contains multiple speakers and diverse recording conditions, making it valuable for examining model generalization across voices and environments. As such, results on \stresspresso should be recognized as complementary evidence of broader generalization beyond \stressds. We encourage readers to listen to examples of both datasets in the project page.

Overall, we define each dataset as $\mathcal{D}=\{x_1,...,x_n\}$ and each sample $x\in\mathcal{D}$ consists of: 
$x=(a, t, s, A, l)$,
where: (i) $t$ is the transcription; 
(ii) $s$ is a binary vector of the stressed words of the transcription $t$, where $1$ marks a word as stressed; (iii) $A$ is a set of two possible interpretations of the transcription $t$, both possible given a different stress pattern; (iv)
$l\in A$ marks the correct interpretation; and finally, (v) $a$ is the speech sample.

\begin{table*}[t!]
\centering
\resizebox{\textwidth}{!}{
\begin{tabular}{llccccc}
\toprule
\multirow{2}{*}{\textbf{Category}} &
\multirow{2}{*}{\textbf{Model}} &
\multirow{2}{*}{\textit{Trans.}} &
\multirow{2}{*}{\textit{Stress}} &
\multirow{2}{*}{\textit{Audio}} &
\multicolumn{2}{c}{\textbf{SSR} $\uparrow$} \\
\cmidrule(lr){6-7}
& & & & & \stressds & \stresspresso \\
\midrule

\multirow{5}{*}{\textit{Text LLM}}
& gpt-4o~\citep{hurst2024gpt} & \cmark & \cmark & \xmark & \textbf{86.2} & \textbf{83.6}  \\
& gpt-4o-mini~\citep{hurst2024gpt} & \cmark & \cmark & \xmark & \underline{79.3} & 80.1  \\
& Llama-3.1-8B-Instruct~\citep{grattafiori2024llama3herdmodels} & \cmark & \cmark & \xmark & 73.3 & \underline{81.1} \\
& Qwen2-7B-Instruct~\citep{yang2024qwen2technicalreport} & \cmark & \cmark & \xmark & 67.8 & 74.2 \\
& Qwen-7B-Chat~\citep{bai2023qwentechnicalreport} & \cmark & \cmark & \xmark & 61.4 & 64.8 \\
\midrule

\multirow{5}{*}{\shortstack[l]{\textit{Cascade} \\ \textit{Pipeline}}}
& \whistress $\rightarrow$ gpt-4o & \xmark & \xmark & \cmark & \textbf{83.4} & \textbf{79.7} \\
& \whistress $\rightarrow$ gpt-4o-mini & \xmark & \xmark & \cmark & \underline{76.1} & 73.2 \\
& \whistress $\rightarrow$ Llama-3.1-8B-Instruct & \xmark & \xmark & \cmark & 65.5 & \underline{76.2} \\
& \whistress $\rightarrow$ Qwen2-7B-Instruct & \xmark & \xmark & \cmark & 65.5 & 69.3 \\
& \whistress $\rightarrow$ Qwen-7B-Chat & \xmark & \xmark & \cmark & 50.9 & 62.3 \\
\midrule

\multirow{9}{*}{\textit{SLM}}
& Gemini-2.5-Pro~\cite{comanici2025gemini25pushingfrontier} & \xmark & \xmark & \cmark & \underline{77.5} & \underline{72.7} \\
& gpt-4o-audio~\citep{hurst2024gpt} & \xmark & \xmark & \cmark & 68.8 & 64.8\\
& Audio-Flamingo-3~\cite{goel2025audioflamingo3advancing} & \xmark & \xmark & \cmark & 56.8 & 52.9 \\
& Qwen3-Omni-30B-Instruct~\cite{xu2025qwen3omnitechnicalreport} & \xmark & \xmark & \cmark & 64.6 & 64.8 \\
& Qwen2Audio-7B-Instruct~\citep{chu2024qwen2audiotechnicalreport} & \xmark & \xmark & \cmark & 53.2 & 51.4 \\
& SALMONN~\citep{tang2024salmonngenerichearingabilities} & \xmark & \xmark & \cmark & 55.9 & 52.4 \\
& LLaMA-Omni~\citep{fang2025llamaomniseamlessspeechinteraction} & \xmark & \xmark & \cmark & 49.5 & 52.9 \\
& Phi-4-multimodal-instruct~\citep{microsoft2025phi4minitechnicalreportcompact} & \xmark & \xmark & \cmark & 52.2 & 50.4 \\
\noalign{\vskip 1.5pt} \cline{2-7} \noalign{\vskip 1.5pt}
& \sftmodel (ours) & \xmark & \xmark & \cmark & \textbf{86.2} & \textbf{87.6}\\
\midrule

Human & Annotators & \xmark & \xmark & \cmark & \textbf{\underline{92.6 (96.0)}} & \textbf{\underline{89.6 (96.0)}} \\
\bottomrule
\end{tabular}}
\caption{\ssr accuracy performance on \textit{\stressds} and \textit{\stresspresso} with different inputs - ground truth transcriptions, stress labels, and the raw audio. We also report human performance and majority vote of three annotators.\label{tab:reasoning_task}}
\end{table*}

\subsection{Evaluation procedure}
\label{subsec:evaluation_procedure}

Given a sample $x$, we evaluate SLMs' ability to perform stress reasoning and stress detection by constructing task-specific prompts. The full set of evaluation prompts used is detailed in Appendix~\ref{appendix:prompts}.

For the sentence stress reasoning (\ssr), we consider two metric variations: (i) \ssr accuracy; and (ii) open-ended \ssr. For \ssr accuracy, the model is provided with the audio and is instructed to select the most likely meaning from a set of possible interpretations. Performance is measured by accuracy. As prompt adherence varies notably across models, we follow the LLM-as-a-judge~\citep{gu2025surveyllmasajudge} approach, using gpt-$4$o \citep{hurst2024gpt}, to interpret model's output. This judgment is then used to compute the final evaluation metric consistently for all evaluated models. Given a model $\mathcal{M}$, dataset $\mathcal{D}$, a prompt $\mathcal{P}$ and a judge $\mathcal{J}$ the \ssr accuracy is measured as follows:
\begin{equation*}
\text{\ssr}(\mathcal{M}, \mathcal{D}) 
= \frac{1}{{|\mathcal{D}}|}
\sum_{x\in\mathcal{D}} \mathbb{I}\{\mathcal{J}(\mathcal{M}(a, \mathcal{P}(A))) = l\}.
\end{equation*}

Additionally, we evaluate the model's capacity to capture stress-derived ambiguity by formulating an open-ended \ssr task. In this setting, we omit predefined interpretations, then the model's free-form answers are scored on a $1$–$5$ scale by an LLM-judge, where a score of $5$ denotes a perfect capture of the intended meaning and a score of $1$ indicates a completely incorrect response.

For sentence stress detection (\ssd), the model receives both the audio $a$ and the ground truth transcription $t$ and is instructed to identify stressed word(s) in the utterance. Then, a judge interprets the model's output and we compute precision, recall, and F$1$ scores based on the predicted stressed words.

\subsection{Benchmarking results}
Equipped with a method to evaluate sentence stress modeling, we benchmark leading SLMs on \stressds and \stresspresso. We consider Qwen3-Omni-30B-Instruct~\cite{xu2025qwen3omnitechnicalreport} Qwen$2$Audio-$7$B-Instruct  \citep{chu2024qwen2audiotechnicalreport}, Audio-Flamingo-3~\cite{goel2025audioflamingo3advancing}, SALMONN \citep{tang2024salmonngenerichearingabilities}, LLaMA-Omni  \citep{fang2025llamaomniseamlessspeechinteraction},  Phi-$4$-multimodal-instruct \citep{microsoft2025phi4minitechnicalreportcompact}, gpt-$4$o-audio \citep{hurst2024gpt} and Gemini-2.5-Pro~\citep{comanici2025gemini25pushingfrontier}. In addition, we conduct a human study to validate that our core \ssr task is relatively easy and straightforward for human listeners. We randomly sample $100$ samples from each dataset, sufficient to estimate human-level performance. We ask annotators to answer the same multiple-choice questions. Each sample is independently labeled by three annotators. We report both overall accuracy and majority-vote accuracy of three annotators in~\Cref{tab:reasoning_task} (bottom cell). The annotation system and protocol can be viewed in Appendix~\ref{appendix:benchmark}, ~\Cref{fig:human_eval}. 

Results suggest that leading SLMs struggle to infer the intended meaning conveyed through stress patterns, most achieving near-random performance, while Gemini-$2.5$-Pro achieves relatively strong \ssr accuracy results of $77.5$ and $72.7$ on \stressds and \stresspresso respectively. 
In contrast, human annotators achieve near-perfect scores, with $96.0\%$ majority-vote accuracy and overall accuracies of $92.6\%$ on \stressds and $89.6\%$ on \stresspresso.

\begin{figure*}[t!]
    \centering
    \includegraphics[width=1\textwidth]{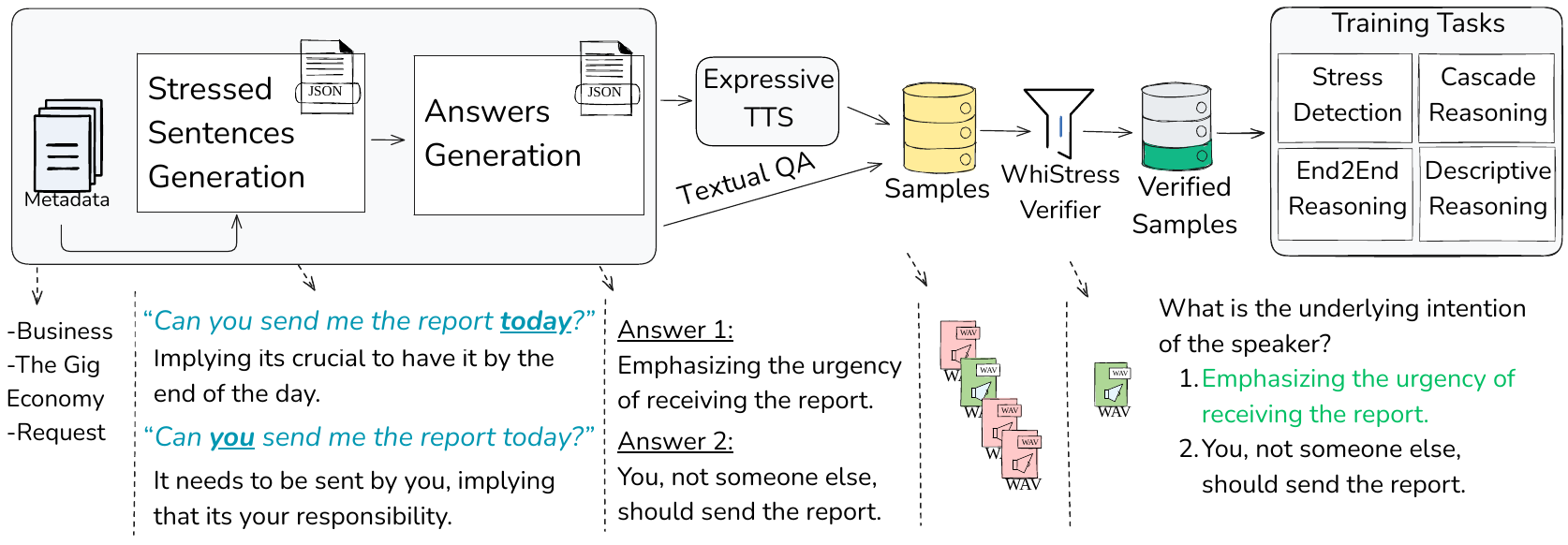}
    \caption{An illustrative example of the synthetic training data generation process. }
    \label{fig:data_generation}
\end{figure*}

\section{Synthetic data generation}
\label{sec:method}

In order to improve model performance on sentence stress modeling and reasoning, we present a synthetic data generation pipeline. With it, we create \traindata, a training set aimed at enhancing performance on sentence stress understanding tasks. The main premise behind this approach is that once data is generated with sufficient diversity and quality, finetuning SLMs on this dataset will generalize to real-world recordings. The data generation methodology is illustrated in \Cref{fig:data_generation} and is divided into four components: (i) Text sample generation; (ii) Stressed speech synthesis; (iii) Stress verification; (iv) Training tasks definition. Overall, the training set amounts to $\sim$$17$K audio samples, of which $\sim$$4.5$K are automatically verified. Additional data statistics are available in Appendix \ref{appendix:training_data}.

\paragraph{Text sample generation.}
We first generate texts, stress patterns, and their interpretations. Note that, as discussed in \Cref{sec:evaluation_benchmark}, not all texts can have different meanings based on emphasis, thus we explicitly create them to that end. The texts are generated through a sequential agentic process, with gpt-$4$o \citep{hurst2024gpt} as the agent, using the CrewAI framework~\citep{crewAI}. 

This sequential process is comprised of two parts: (i) Prompt the agent to create a sentence that can be understood differently according to the stressed words. In order to increase diversity, and avoid repetitions, the instruction is given a domain, topic and sentence type from a list. Following notations from~\Cref{sec:eval_ds}, after this phase we have  two samples with the same transcription: $x_1=(t, s_1, d_1), x_2=(t, s_2, d_2)$ where for $i\in\{1,2\}$, $d_i$ describes the underlying meaning of each interpretation implied by the stress, $s_i$; (ii) As $d$ can be lengthy, we aim to shorten it. Given the input from (i), we prompt the agent to create a set of answers $A$, each summarizing the corresponding interpretation's description $d$. Hence, $h_i\in A$ can seen as a concise version of $d_i$. Then, we end with two samples of the form $x=(t, s, d, A, l)$ where $l\in A$ is the target answer out of two possible answers corresponding to $t$. By that, we can ask a model what is the speaker's intention while providing two feasible answers. Prompts and metadata can be seen in Appendix~\ref{appendix:data_generation_textual_samples}.

\paragraph{Stressed speech synthesis.}
Given text samples with known words to stress to convey a desired meaning, we use the OpenAI text-to-speech~\citep{OpenAITTS} to generate stressed speech. We find that marking the stressed words with enclosing asterisks leads to them being synthesized as stressed. Our preliminary results suggest, this approach leads to more natural voice compared to editing prosodic features directly. For each text stress pattern, we generate two audio samples using randomly selected male or female speakers. This results in four audio samples per transcription $t$: two reflecting one stress pattern and two reflecting a different one. Finally, each sample is of the form $x=(a, t, s, d, A, l)$ where $a$ marks the synthesized speech.

\paragraph{Stress verification.} 
Despite using expressive TTS, our studies reveal frequent stress errors, such as misplaced or missing emphasis. To address this, we use \whistress \citep{whistress}, which predicts transcriptions and stressed words, allowing us to filter out incorrectly stressed samples. This yields a higher-quality subset (see Appendix \ref{appendix:training_data_validation}) and improves performance, particularly under a curriculum setup (see \Cref{sec:ablation}).

\paragraph{Training task definition.}
\label{subsubsection:training_tasks}
We aim to improve SLM performance on \ssr and \ssd. To this end, each sample $x=(a, t, s, d, A, l)$ is utilized in four tasks, with a corresponding prompt: (i) \emph{Sentence stress detection}: The model is requested to identify the stressed words $s$, given the ground truth transcription $t$; (ii) \emph{End-to-end reasoning}: The model is asked to choose the most likely underlying meaning out of $A$ according to the audio, directly responding the correct answer $l\in A$; (iii) \emph{Elaborate reasoning}: Similar to the above, this task aims to train the model on \ssr. However, the model is instructed to first elaborate on the interpretation's meaning and then choose the most likely answer. In practice, we use the description $d$ as the model's elaboration prefix before the final answer $l$; (iv) \emph{Cascade reasoning}: This task targets both \ssd and \ssr by requesting the model to output the transcription $t$ with the emphasized words $s$ and then output the correct answer $l$. We hypothesize that this variation allows the model to better connect sentence stress and implied underlying meaning. Notice, we compute the loss with regards to the entire answer (including elaborations). The training prompts are given in Appendix~\ref{appendix:training_prompts}. 

\begin{table*}[t!]
\small
\centering
\resizebox{\textwidth}{!}{
\begin{tabular}{lccc|ccc|ccc}
\toprule
\multirow{4}{*}{\textbf{Model}} 
& \multicolumn{9}{c}{\textbf{SSD} $\uparrow$} \\
\cmidrule(lr){2-10}
& \multicolumn{3}{c}{\textbf{Expresso}} 
& \multicolumn{3}{c}{\textbf{\stressds}} 
& \multicolumn{3}{c}{\textbf{\stresspresso}} \\
\cmidrule(lr){2-4} \cmidrule(lr){5-7} \cmidrule(lr){8-10}
& \textit{P.} & \textit{R.} & \textit{F1} 
& \textit{P.} & \textit{R.} & \textit{F1} 
& \textit{P.} & \textit{R.} & \textit{F1} \\
\midrule
\whistress & \textbf{\underline{57.3}} & \textbf{\underline{86.3}} & \textbf{\underline{68.9}} 
           & \textbf{\underline{88.5}} & \textbf{\underline{88.1}} & \textbf{\underline{88.3}} 
           & \textbf{\underline{74.0}} & \textbf{\underline{95.7}} & \textbf{\underline{83.5}} \\
\midrule
Gemini-2.5-Pro & 26.1 & \textbf{82.1} & \underline{39.6} 
               & \underline{35.9} & 74.9 & \underline{48.5} 
               & \underline{27.5} & 77.8 & \underline{40.7} \\
gpt-4o-audio & 23.6 & 66.1 & 34.7 
             & 32.4 & \underline{79.7} & 46.1 
             & 23.8 & \underline{81.6} & 36.9 \\
Audio-Flamingo-3 & 19.5 & \underline{75.2} & 31.0 
                 & 26.3 & 72.6 & 38.7 
                 & 18.9 & 76.8 & 30.4 \\
Qwen3-Omni-30B-Instruct & 27.0 & 69.2 & 38.9 
                        & 34.2 & 71.2 & 46.2 
                        & 27.7 & 69.8 & 39.7 \\
Qwen2Audio-7B-Instruct & \underline{34.2} & 30.6 & 32.3 
                       & 22.9 & 59.7 & 33.1 
                       & 17.1 & 50.0 & 25.5 \\
SALMONN & 13.2 & 45.5 & 20.5 
        & 20.6 & 41.6 & 27.5 
        & 14.5 & 55.1 & 23.0 \\
LLaMA-Omni & 18.7 & 58.2 & 28.3 
           & 25.5 & 48.7 & 33.5 
           & 17.0 & 40.0 & 23.8 \\
Phi-4-multimodal-instruct & 22.5 & 37.5 & 28.2 
                          & 20.7 & 36.5 & 26.5 
                          & 18.6 & 42.9 & 26.0 \\
\midrule
\sftmodel (ours) & \textbf{51.8} & 68.6 & \textbf{59.1} 
                 & \textbf{89.4} & \textbf{84.5} & \textbf{86.9} 
                 & \textbf{77.1} & \textbf{84.4} & \textbf{80.6} \\
\bottomrule
\end{tabular}}
\caption{Sentence stress detection (\ssd) performance across Expresso, \stressds, and \stresspresso datasets. We compare the results of \sftmodel to other SLMs, and also to \whistress, which is trained solely for \ssd. We report Precision (P), Recall (R) and F1.}
\label{tab:stress_detection}
\end{table*}

\section{Experiments and results}
\label{sec:results}

We study the efficacy of our synthetic training data, \traindata, on \ssr and \ssd. We further conduct analysis to ablate the impact of pipeline components - stress verifier, trained encoder and training tasks.

We finetune Qwen2Audio-7B-Instruct \citep{chu2024qwen2audiotechnicalreport} using LoRA adapters \citep{hu2021loralowrankadaptationlarge} on the query and value projections with the synthetic training dataset generated by our proposed pipeline. To prevent overfitting to stress-focused tasks, we also add samples from the original tasks which the base model was trained on, namely, LibriLight \citep{Kahn_2020} for automatic-speech-recognition (ASR) and MELD \citep{poria2019meldmultimodalmultipartydataset} for speech-emotion-recognition (SER). We ensure that the total audio duration of these auxiliary tasks approximately matches our verified training subset.

During training, we employ a staged training approach, where the model is first finetuned on the full \traindata (both verified and unverified) for one epoch, then finetuned on a high-quality subset for another. We find this two-stage strategy effective in balancing the performance on both \ssr and \ssd tasks. Hyperparameters and implementation details are reported in Appendix \ref{appendix:training_procedure}.

\subsection{Sentence stress detection}
One may argue that an essential prerequisite for understanding the meaning conveyed by sentence stress is the ability to accurately detect it. For this purpose, we start by evaluating the ability of SLMs on the \ssd task. We evaluate models on \stressds, \stresspresso and Expresso~\citep{nguyen2023expressobenchmarkanalysisdiscrete} benchmarks. To align with previous work \citep{whistress, deseyssel2024emphassessprosodicbenchmark} we use samples with at least one stressed word of speakers ex$01$ and ex$02$. Results in~\Cref{tab:stress_detection} show that current SLMs struggle with detecting stress, reaching at most F1 of 48.5. In contrast, \sftmodel achieves F1 scores above 80.6 on our proposed datasets, marking a substantial improvement.

\subsection{Sentence stress reasoning}
We evaluate \ssr performance across different input settings in both speech-aware and text-only LMs. Since SLMs are an extension of text LMs, we analyze the scenario where both transcription and stressed words are given as input. We consider three primary settings: (i) an oracle configuration in which both ground truth transcription and stressed words are given to an LM; (ii) a cascade pipeline in which these inputs are predicted by \whistress; (iii) a fully end-to-end setting, where SLMs receive only the raw audio and directly predict the underlying meaning. Results are summarized in~\Cref{tab:reasoning_task}, and additional settings tested are available in~\Cref{tab:reasoning_task_ext}.

The proposed model demonstrates strong results on \ssr, exceeding evaluated SLMs and cascade models that receive audio only as input, $86.2, 87.6$ vs. $83.4, 79.7$ on \stressds and \stresspresso respectively, when comparing to WhiStress followed by gpt-4o. This demonstrates a possible benefit to the direct approach vs a cascade, even against a much larger and stronger text model. It further shows the efficacy of the data synthesis pipeline.

In assessing sentence stress understanding in text LLMs, we note, a somewhat expected drop for cascade versions against the oracle stress labels on \ssr, for instance $73.3$ vs. $65.5$ for the Llama$3.1$ version. This highlights the error propagation in cascade approaches. Moreover, the results indicate that performance of text LLMs tends to correlate with the model’s overall language capabilities—as reflected by their scores on standard text benchmarks—where open-source 7B models lag behind proprietary models like gpt-4o-mini and gpt-4o.

Despite error propagation, the cascade approach outperforms current end-to-end SLMs (with the exception of our approach). This implies that stress information remains underutilized when derived directly from raw audio, thereby highlighting the effectiveness of our pipeline in extracting and leveraging prosodic cues for reasoning.

Finally, while Gemini-2.5-Pro performs worse than \sftmodel on both datasets, it is the only SLM to surpass $70$ in \ssr accuracy. We attribute this to its high \ssd recall (\Cref{tab:stress_detection}), which, combined with two possible answers given as context, may help the model identify the likely stressed words. Moreover, as shown in~\Cref{tab:reasoning_task_ext}, Gemini notably outperforms other LMs when given ground-truth stress labels, showing its strong text reasoning ability.

\subsection{Open-ended sentence stress reasoning}
\begin{figure}[t!]
    \centering
    \includegraphics[width=\columnwidth]{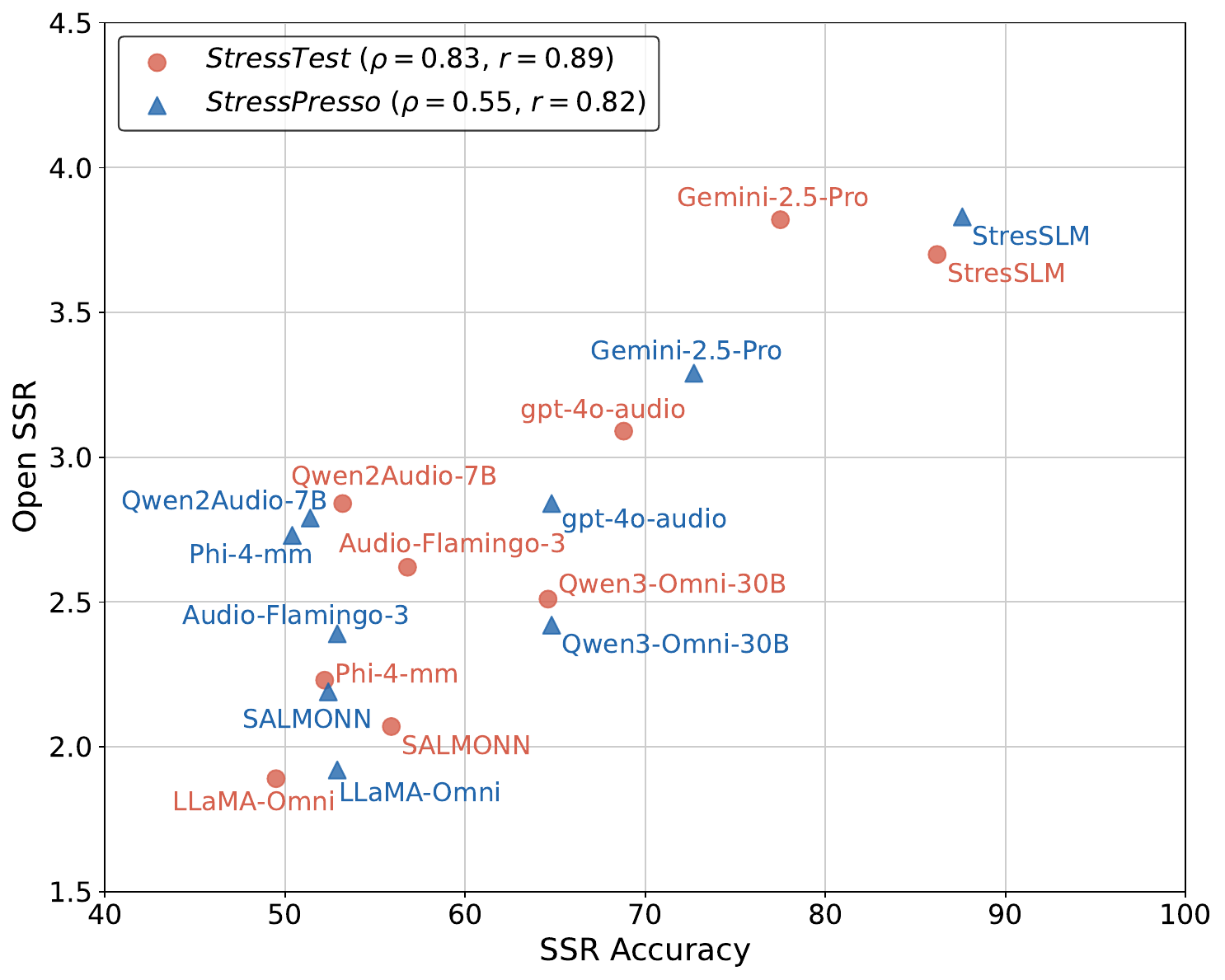}
    \caption{Relationship between \ssr accuracy and open-ended \ssr across SLMs. Spearman and Pearson correlation coefficients are denoted by $\rho$ and $r$ respectively.}
    \label{fig:open_closed_ssr}
\end{figure}

We explore the ability of SLMs to handle stress-derived ambiguity using an open-ended \ssr evaluation. The results are presented in Table~\ref{tab:open_ssr} in Appendix~\ref{appendix::open_ssr} and illustrated in Figure~\ref{fig:open_closed_ssr}. Overall, these findings are consistent with \ssr accuracy trends and suggest that most existing SLMs struggle to capture stress-related meaning in this free-form setting. In contrast, \sftmodel demonstrates stronger generalization, more effectively capturing the variability of stress meanings across different acoustic conditions. Importantly, the results indicate that our method enables \sftmodel to reason about stress-implied intent, rather than merely performing discriminative selection.

Additionally, we examine the consistency between binary-choice QA and open-ended evaluations. The results indicate a high correlation between \ssr accuracy and open-ended \ssr performance, suggesting that \ssr accuracy can serve as an efficient proxy for general stress reasoning in capable SLMs. However, as expected, this relationship becomes less reliable for lower-performing models, where near-random accuracy introduces higher variability and weakens the correlation between the two metrics.

\begin{table}[t!]
\centering
\resizebox{\columnwidth}{!}{
\begin{tabular}{l|cc|c}
\toprule
\multirow{3}{*}{\textbf{Model}} & \multicolumn{2}{c|}{\textbf{ASR (WER) $\downarrow$}} & \textbf{SER} $\uparrow$ \\
\cmidrule(lr){2-3} \cmidrule(lr){4-4}
& \textbf{LS} & \textbf{CV} & \textbf{MELD} \\
& \textit{clean} / \textit{other} & \textit{test} & \textit{test} \\
\midrule
Qwen2Audio-7B & \textbf{1.73 / 4.01} & \textbf{8.72} & 54.6 \\
Qwen2Audio-7B-Inst. & 2.31 / 4.92 & 11.49 & 26.4 \\
\sftmodel (ours) & 2.29 / 4.59 & 10.49 & \textbf{57.6} \\
\bottomrule
\end{tabular}}
\caption{Results on core speech tasks. ASR shown on test split of LibriSpeech (LS) and CommonVoice (CV).}
\label{tab:librispeech_wer}
\end{table}

\subsection{\stressds vs. \stresspresso}
Overall, results on both datasets show consistent trends, with \sftmodel leading on \ssd ($88.3$ and $83.5$) and \ssr accuracy ($86.2$ and $87.6$), and maintains high performance and robustness in open-ended reasoning ($3.70$ and $3.83$) on \stressds and \stresspresso, respectively. 
While Gemini-2.5-Pro demonstrates impressive capabilities achieving a high score of $3.82$ on the controlled \stressds, its performance drops significantly to $3.29$ on \stresspresso. This disparity indicates a lack of robustness in diverse acoustic settings. In contrast, all open-source models perform near-randomly in binary selection and consistently yield results that reflect poor or ambiguous reasoning in open-ended \ssr. This demonstrates that finetuning on our generated data not only enhances sentence stress understanding on \stressds, but also enables the model to generalize to additional speakers and recording conditions in \stresspresso.

\subsection{Effect on original tasks}
To assess whether our approach introduces trade-offs with existing capabilities, we evaluate \sftmodel on original tasks used to train the base model. Specifically, we test ASR, over LibriSpeech test sets and CommonVoice, and SER using MELD. This analyzes if our stress-focused training interferes with the SLM's abilities on basic speech understanding tasks. Results are reported in~\Cref{tab:librispeech_wer}. We observe no degradation in ASR or SER compared to the Qwen2Audio-$7$B-Instruct model, indicating that the model retains its speech recognition ability. These findings suggest that our multitask training setup does not inherently conflict with the original objectives, demonstrating the potential to enrich SLMs with sentence stress reasoning capabilities, without compromising on established tasks.

\subsection{Ablation study}
\label{sec:ablation}
We analyze the impact of our design choices in training \sftmodel, on the performance on \stressds. Models are trained on \traindata without the ASR and SER rehearsal data. Additionally, all models are trained under the same conditions with a compute budget equivalent to 5 epochs on the verified data subset, the number of steps is smaller when removing training tasks. Further results on \stresspresso, in Appendix~\ref{appendix:ablation}, show similar trends.

\paragraph{Staged training introduces balance.} We first evaluate the efficacy of the \whistress verifier. We train models on all stress-related tasks using a frozen speech encoder and assess performance under different training strategies. Results, presented in~\Cref{tab:verifier_effectiveness}, show that training on the verified subset improves performance on \ssr, suggesting that the data selected by \whistress is beneficial for stress reasoning. However, this comes at the cost of reduced performance on the \ssd task, likely due to reduced data diversity or quantity. In contrast, the staged training approach, in which the model is first trained on the full dataset and then fine-tuned on the filtered subset, achieves a better balance, improving \ssd performance while slightly reducing \ssr. 

\begin{table}[t!]
    \centering
    \resizebox{\columnwidth}{!}{
    \begin{tabular}{cccccc}
    \toprule
    \multirow{2}{*}{\textbf{Verifier}} & \multirow{2}{*}{\textbf{\# Samples}} & \multicolumn{3}{c}{\textbf{SSD $\uparrow$}} & \textbf{SSR} $\uparrow$ \\
    \cmidrule(r){3-5} \cmidrule(r){6-6}
      &  & \textit{P.} & \textit{R.} & \textit{F1} & \textit{Acc.} \\
    \midrule
    \cmark & $\sim$4K & 87.3 & 76.3 & 81.4 & \textbf{79.3}\\
    \xmark & $\sim$17K & \underline{87.4} & \underline{81.9} & \underline{84.5} & 76.6 \\
    \xmark $\rightarrow$ \cmark & 17K$\rightarrow$4K & \textbf{88.3} & \textbf{83.7} & \textbf{85.9} & \underline{78.4}\\
    \bottomrule
    \end{tabular}}
    \caption{Effect of using \whistress verifier on model performance on \stressds benchmark. We report Precision (P), Recall (R), F1, and Accuracy (Acc).}
\label{tab:verifier_effectiveness}  
\end{table}

\paragraph{Speech encoder captures stress.} We assess whether stress related information must be explicitly extracted from the audio encoder of \sftmodel, or if the model can rely solely on frozen speech representations from the original model. As shown in~\Cref{tab:ablation}, the model benefits from training the encoder on both \ssr and \ssd tasks. This finding is consistent with previous works \citep{whistress,pasad2022layerwiseanalysisselfsupervisedspeech} which demonstrated that prosodic features are encoded in different layers of speech representation models, thus, finetuning the encoder enables extracting this stress-related information. Moreover, the results indicate that SLMs finetuned on the \ssd task can outperform existing stress detection models such as \whistress. As shown in~\Cref{tab:stress_detection}, \whistress achieves F$1$ scores of $88.3$ on \stressds, while \sftmodel that was trained solely on stress detection-oriented tasks achieves higher F$1$ scores of $90.5$ (\Cref{tab:ablation}).

\paragraph{Training tasks balance each other.} We analyze the effect of the different tasks introduced in~\Cref{subsubsection:training_tasks}. To do so, we conduct an ablation by removing one training task at a time (always retaining the end-to-end reasoning task). Results shown in~\Cref{tab:ablation}, show that no combination yields the best performance for all metrics. However, including all tasks produces second-best results for both \ssr and \ssd, suggesting that a diverse task mixture leads to more balanced training. Interestingly, removing the elaborated explanation task results in a substantial drop in \ssr performance, while improving \ssd results. This may be due to \ssd-oriented data becoming more dominant in the training data, indicating a trade-off in task emphasis.

\begin{table}[t!]
    \centering
    \resizebox{\columnwidth}{!}{
    \begin{tabular}{cccc|cccc}
    \toprule
    \textit{Train} & \textit{Stress} & \textit{Cas.} & \textit{Elab.} & \multicolumn{3}{c}{\textbf{SSD} $\uparrow$}  & \multicolumn{1}{c}{\textbf{SSR} $\uparrow$} \\
    \cmidrule(r){5-7} \cmidrule(r){8-8}
    \textit{Enc.} & \textit{Det.} & \textit{Reas.} & \textit{Reas.} & \textit{P.} & \textit{R.} & \textit{F1} & \textit{Acc.}  \\
    \midrule
    \xmark & \cmark & \cmark & \cmark & 88.3 & 83.7 & 85.9 & 78.4 \\
    \cmidrule(r){1-8}
    \cmark & \xmark & \cmark & \cmark & 45.5 & 86.3 & 59.9 & \textbf{84.4} \\
    \cmark & \cmark & \xmark & \cmark & 90.4 & 83.3 & 86.7 & 82.5 \\
    \cmark & \cmark & \cmark & \xmark & \textbf{94.4} & \textbf{87.0} & \textbf{90.5} & 78.4 \\
    \cmidrule(r){1-8}
    \cmark & \cmark & \cmark & \cmark & \underline{92.4} & \underline{86.3} & \underline{89.3} & \underline{83.0} \\
    \bottomrule
    \end{tabular}}
    \caption{Ablating different training choices of \sftmodel on stress modeling capabilities, using staged-training. We report Precision (P), Recall (R), F1, and Accuracy (Acc) for encoder training, stress detection, cascade reasoning, and elaborated reasoning.}
    \label{tab:ablation}
\end{table}
\section{Related work}
\label{sec:related_work}

\paragraph{Sentence stress modeling.} Existing work define sentence stress detection as word-level binary classification. Approaches aim to directly detect stress from speech \citep{ProsodicStressRevisited, mishra12_interspeech} or also include grammar and contextual information \citep{Lin2020JointDO, koreanarticle}. Many modern methods use speech representations from pre-trained models \citep{deseyssel2024emphassessprosodicbenchmark, whistress}. Another effort is developing expressive TTS models that can generate emphasized speech \citep{stephenson2022bertpredictcontrastivefocus}. Other work explore if text LLMs understand the meaning of emphasized words in discourse \citep{lin-lee-2024-llms}.

\paragraph{Speech-aware LMs.} Recently, integrating speech and audio to LLMs has demonstrated impressive abilities on speech tasks \citep{arora2025landscapespokenlanguagemodels}. These models follow a similar approach - use a pretrained LLM, encode speech to a latent representation with a pretrained encoder, and project it to the LLM embedding space. 
Qwen2Audio use a Whisper encoder \citep{radford2022robustspeechrecognitionlargescale} and Qwen-7B \citep{bai2023qwentechnicalreport}. Other models use additional encoders \citep{tang2024salmonngenerichearingabilities}, introduce the ability to generate speech \citep{fang2025llamaomniseamlessspeechinteraction, xie2024miniomnilanguagemodelshear}, or add modalities \citep{microsoft2025phi4minitechnicalreportcompact, xu2025qwen3omnitechnicalreport}.

\paragraph{Related benchmarks.} SLM evaluation spans diverse tasks, e.g. audio question answering \citep{lipping2022clothoaqacrowdsourceddatasetaudio}, speech-to-text-translation \citep{wang2020covost2massivelymultilingual}. Several evaluation suites cover many tasks, including prosodic elements \citep{yang2024airbenchbenchmarkinglargeaudiolanguage, wang2025audiobenchuniversalbenchmarkaudio, yang2021superbspeechprocessinguniversal, maimon2025salmonsuiteacousticlanguage, ma2025c3bilingualbenchmarkspoken}. 
For stress detection, a few benchmarks have been proposed, relying on synthetic data or existing emotion datasets \citep{whistress, deseyssel2024emphassessprosodicbenchmark}. Another work includes crowd-sourcing to collect stress annotations in spoken utterances \citep{morrison2023crowdsourcedautomaticspeechprominence}. Recently, \citet{deseyssel2024emphassessprosodicbenchmark} introduced a framework to assess if spoken translation models preserve stress. Despite this, there remains a lack of benchmarks to evaluate models' ability to reason about meaning of spoken stress.

\section{Conclusion}
\label{sec:conclusion}
We introduce \stressds, a benchmark for evaluating stress understanding in SLMs, considering two tasks: Sentence Stress Reasoning and Detection. These tasks highlight the important yet underexplored role of stress placement in shaping meaning in spoken language, particularly in the context of speech-aware LMs, exposing a clear gap in existing models. We then propose a novel automatic data synthesis pipeline tailored to stress modeling, by which creating \traindata. Finetuning on \traindata allows our model, \sftmodel, to significantly surpass existing SLMs on both \ssr and \ssd, while preserving performance on core tasks. 

\section*{Limitations}
Despite these contributions, our work is currently limited to English and limited speakers. How well stress-based reasoning generalizes across languages, accents and in conversational settings remains an open question. Our findings highlight the value of prosody in speech-language modeling and suggest promising directions for future work in speech understanding and generation that better reflect the nuances of human communication.

\paragraph{Acknowledgments.} This research work was supported by ISF grant 2049/22.

\bibliography{bib}
\clearpage
\appendix

\section{Benchmark information}
\label{appendix:benchmark}
\stressds comprises $101$ unique texts, each recorded with at least two distinct sentence stress patterns, yielding different underlying interpretations of the same utterance. Specifically, $85$ sentences have $2$ different interpretations, while $16$ have $3$. Of the interpretations - $170$ have a single stressed word, $43$ contain $2$ stressed words and $5$ include $3$ distinct stressed words. In total, the number of audio samples is $218$. All audio recordings are sampled of $48$kHz. To match SLMs requirements, audio recordings were down-sampled to $16$kHz. All recordings were approved by the Ethics committee at the university the dataset were recorded, where the professional actor get paid more than $5$ times the base salary.   

\stresspresso contains $202$ audio samples, each with a unique stress pattern, derived from $96$ distinct texts. Of these, $113$ are female recordings ($51$ from speaker ex02 and $62$ from ex04) and $89$ are male recordings ($31$ from ex01 and $58$ from ex03).

\paragraph{Recording information.} All audio samples were recorded using Shure-SM$7$ microphone over RME-$800$ sound card in a professional and acoustically treated studio. All samples were recorded in English by a professional actor, who got paid more than $4$ times the minimum wage.  

\begin{figure}[h!]
    \centering    \includegraphics{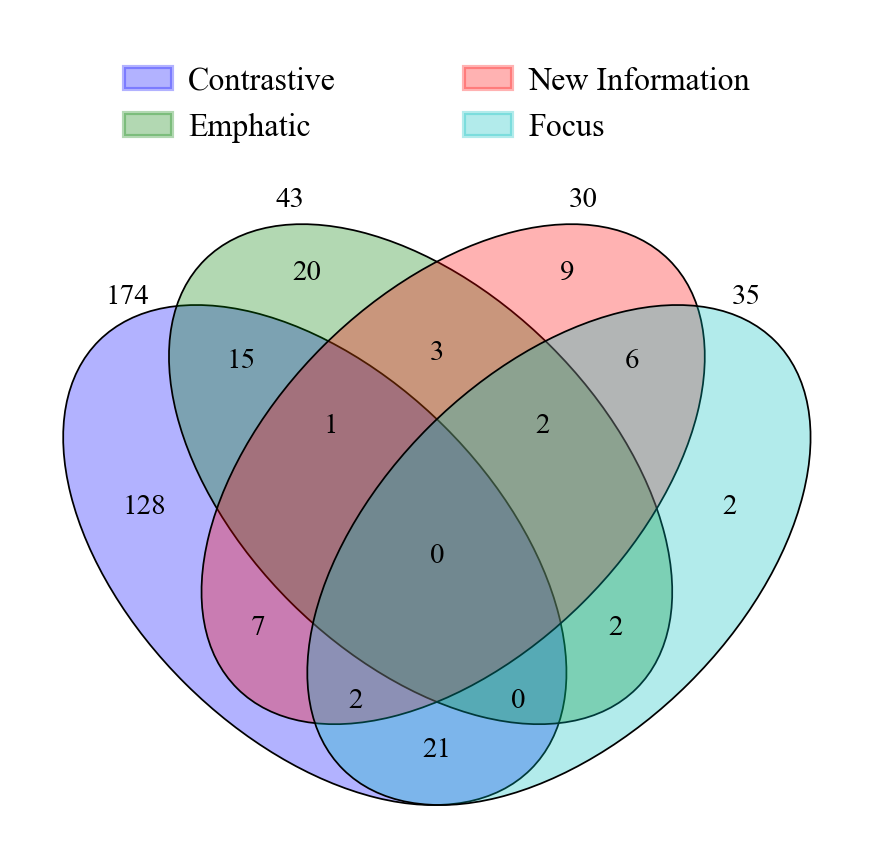}
    \caption{Categorization of sentence stress types in \stressds.}
    \label{fig:stress_ven}
\end{figure}

\paragraph{Types of sentence stress in \stressds.} Based on the stress types presented in Section~\ref{background}, we coarsely categorize and illustrate the different types of sentence stress represented in the \stressds samples in~\Cref{fig:stress_ven}. As definitions of sentence stress are fluid and not strictly defined, categorization inaccuracies may be present.

\begin{figure*}[t!]
    \centering
    \includegraphics[width=1\textwidth]{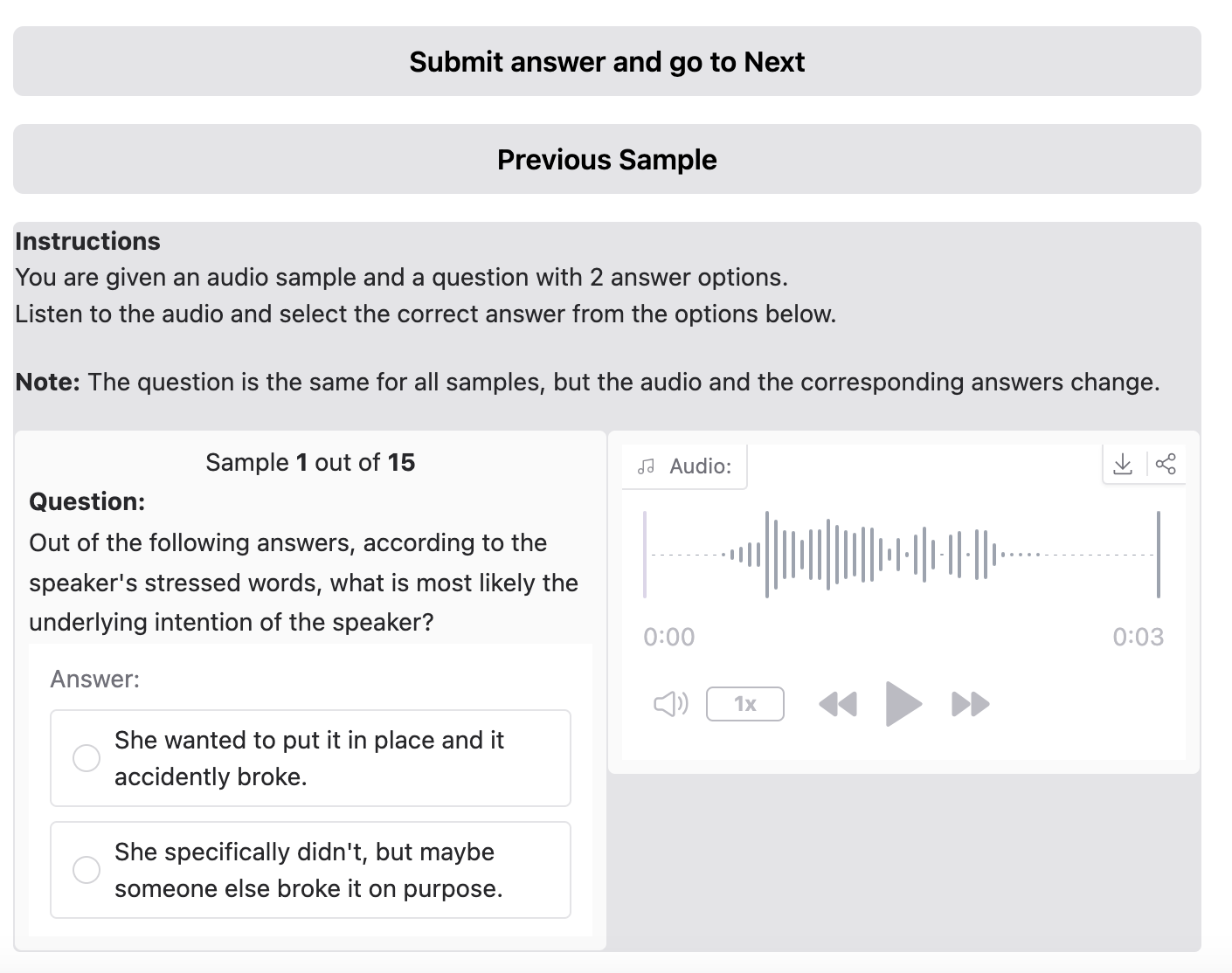}
    \caption{Human evaluation annotation view.}
    \label{fig:human_eval}
\end{figure*}

\paragraph{Human evaluation.} Human annotators where requested to fill forms with 15 samples to evaluate. The form is illustrated in Figure~\ref{fig:human_eval}.

\section{Training data and training procedure}
\subsection{\traindata information}
\label{appendix:training_data}
Table~\ref{tab:dataset_stats} features statistics about the synthetically generated dataset before and after the use of the \whistress verifier. The number of samples in the table is multiplied by the number of task templates presented in Section~\ref{sec:method} resulting in $\sim$$17$K samples.

\begin{table}[t!]
\centering
\begin{tabular}{lcc}
\toprule
\textbf{Statistic} & \textbf{Full} & \textbf{Verified} \\
\midrule
\# Samples (audio)              & 4400 & 1311 \\
\# Unique Interpretations        & 2200 & 931  \\
\# Unique Transcriptions         & 1100 & 731  \\
\# Trans. with $\geq$2 Interp. & 1100 & 200  \\
\midrule
\textbf{Gender Distribution} & & \\
\midrule
Female                  & 2122 & 597  \\
Male                    & 2278 & 714  \\
\bottomrule
\end{tabular}
\caption{\traindata statistics before and after verifying with \whistress.}
\label{tab:dataset_stats}
\end{table}

\subsection{\traindata data validation}
\label{appendix:training_data_validation}
We conducted a small-scale human study to validate the synthetic data. We asses the TTS model's ability to generate stressed speech, alignment between stressed speech and textual explanations and the WhiStress verifier effectiveness.

We use SSD as a proxy for measuring stressed TTS quality, and SSR for alignment of the text and TTS generated audio. We randomly sampled 20 recordings from the WhiStress verified subset, and 20 samples from the non-verified subset. Then, each synthetic sample was annotated by 3 different annotators, completing both SSD and SSR tasks. The SSD and SSR are calculated based on the desired TTS stress label vs the annotator stress marking as the target answer. For SSR, we report accuracy across all annotated samples, 60 per subset, as well as majority vote accuracy between the 3 annotators on the 20 samples of each subset. In the SSD task, we report mean scores for 3 annotators where each score was calculated on 20 samples. Results are summarized in~\Cref{tab:verified_vs_nonverified}.

\begin{table}[t!]
\centering
\resizebox{\columnwidth}{!}{
\begin{tabular}{l|ccc|c}
\toprule
\multirow{2}{*}{\textbf{Subset}} & \multicolumn{3}{c|}{\textbf{SSD} $\uparrow$} & \textbf{SSR} $\uparrow$ \\
\cmidrule(lr){2-4} \cmidrule(lr){5-5}
& \textit{P.} & \textit{R.} & \textit{F1} & \textit{Acc.} \\
\midrule
Verified     & \textbf{72.30} & \textbf{86.67} & \textbf{78.81} & \textbf{83.33} (\textbf{85.0}) \\
Non-verified & 47.12 & 55.35 & 50.75 & 63.33 (70.0) \\
\bottomrule
\end{tabular}}
\caption{Human evaluation results comparing verified and non-verified \traindata subsets. SSR accuracy is reported with majority vote in parentheses.}
\label{tab:verified_vs_nonverified}
\end{table}

\paragraph{Verified data captures stress better.} Results suggest that the verified subset adheres to the stressed synthetic speech more than the non-verified subset, achieving F1 score of 78.81\% v.s 50.75\% respectively. Additionally, in terms of alignment between the stress and the underlying intention presented in the textual answers of the SSR task, the verified subset results outperform those of the non-verified, with 85\% compared to 70\%. This shows the efficacy of the WhiStress verifier to capture stress and create a higher quality subset of the training data.

\paragraph{Non-verified data contains meaningful signals.} Results indicate that valuable positive signal is present in the non-verified training data for detecting and reasoning about stress as well. This validates the ability of the training data to address our proposed tasks and supports our motivation to first train on large, noisy, data, and then continue training on a smaller but higher quality filtered data.

\subsection{Training procedure}
\label{appendix:training_procedure}
During training, gradients are computed only over the model's final answers. Hyperparameters for all models were chosen using grid search, taking the model to reach the best results on the validation set.

\begin{table}[t!]
\centering
\resizebox{\columnwidth}{!}{
\begin{tabular}{lc}
\toprule
\textbf{Hyperparameter} & \textbf{Value} \\
\midrule
Learning Rate & 7e-5 \\
Batch Size (per device) & 8 \\
Gradient Accumulation Steps & 2 \\
\midrule
\multicolumn{2}{l}{\textbf{LoRA Configuration} \citep{hu2021loralowrankadaptationlarge}} \\
\midrule
Rank $r$ & 16 \\
LoRA $\alpha$ & 32 \\
Use rslora \citep{kalajdzievski2023rankstabilizationscalingfactor} & True \\
Target Modules & \texttt{q,v proj} \\
LoRA Dropout & 0.1 \\
\bottomrule
\end{tabular}}
\caption{Training hyperparameters.}
\label{tab:training_hyperparams}
\end{table}

\paragraph{Final model.} The final model was trained for 1595 steps using a cosine learning rate scheduler with a warm-up ratio of 5\%. In the first 1261 steps, we use the entire data set (stage 1), followed by continued training in the verified subset for the remaining steps (stage 2), while preserving the internal state of the scheduler. The model was finetuned on a single NVIDIA L40S GPU. Training hyperparameters are summarized in Table~\ref{tab:training_hyperparams}.

\paragraph{Ablation models.} When trained on the verified only and the full datasets the checkpoint reaching the best performance on the validation set is chosen, while in the staged training the best checkpoint in the second stage is considered.


\section{Additional results}
\subsection{Extended inputs}

\begin{table*}[t!]
\centering
\begin{tabular}{llllcc}
\toprule
\textbf{Model} & \multicolumn{3}{c}{\textbf{Input}} & \multicolumn{2}{c}{\textbf{\ssr} $\uparrow$} \\
 & \textit{Trans.} & \textit{Stress} & \textit{Audio} & \stressds & \stresspresso \\
\midrule 
\textit{SLM}\\
\midrule
Gemini-2.5-Pro & \cmark & \cmark & \cmark & \textbf{98.1} & \textbf{96.5} \\
gpt-4o-audio & \cmark & \cmark & \cmark & \underline{87.1} & 85.6 \\
Audio-Flamingo-3 & \cmark & \cmark & \cmark & 66.0 & 75.2 \\
Qwen3-Omni-30B-Instruct & \cmark & \cmark & \cmark & 84.8 & 89.6 \\
Qwen2-Audio-7B-Instruct & \cmark & \cmark & \cmark & 60.5 & 62.3 \\
SALMONN & \cmark & \cmark & \cmark & 59.6 & 62.8 \\
LLaMA-Omni & \cmark & \cmark & \cmark & 70.1 & 71.2 \\
Phi-4-multimodal-instruct & \cmark & \cmark & \cmark & 55.5 & 56.9 \\
\sftmodel (ours) & \cmark & \cmark & \cmark & 83.4 & \underline{92.5} \\
\midrule
Gemini-2.5-Pro & \xmark & \cmark & \cmark & \textbf{97.2} & \textbf{94.5} \\
gpt-4o-audio & \xmark & \cmark & \cmark & 83.9 & 85.0 \\
Audio-Flamingo-3 & \xmark & \cmark & \cmark & 64.6 & 68.3 \\
Qwen3-Omni-30B-Instruct & \xmark & \cmark & \cmark & \underline{84.4} & 87.1 \\
Qwen2-Audio-7B-Instruct & \xmark & \cmark & \cmark & 59.1 & 65.3 \\
SALMONN & \xmark & \cmark & \cmark & 66.5 & 69.3 \\
LLaMA-Omni & \xmark & \cmark & \cmark & 62.3 & 67.8 \\
Phi-4-multimodal-instruct & \xmark & \cmark & \cmark & 51.3 & 55.4 \\
\sftmodel (ours) & \xmark & \cmark & \cmark & 78.8 & \underline{92.0} \\
\midrule
Human & \xmark & \xmark & \cmark & \textbf{\underline{92.6 (96.0)}} & \textbf{\underline{89.6 (96.0)}} \\
\bottomrule
\end{tabular}
\caption{Performance on SSR the task of \textit{\stressds} using different input configurations. The input column demonstrates the information provided to the model in addition to our task's question. This extends upon table \ref{tab:reasoning_task}.}
\label{tab:reasoning_task_ext}
\end{table*}

In this section we analyze the impact of performance on providing SLMs with the audio as well as ground truth stress and optionally the ground truth transcription. This can be seen as extending table~\ref{tab:reasoning_task}, helping understand if the SLMs utilize the audio input, and how much of the error has to do with transcription performance. The prompts used for this analysis are provided in~\Cref{subsec:reasoning_analysis_prompts} and the results are provided in Table \ref{tab:reasoning_task_ext}. 

We test two scenarios: (i) An oracle provides the SLM with ground truth transcription and ground truth stress in addition to the provided audio; and, (ii) An oracle provides solely the ground truth stress while relying on the ability of the SLM to recognize the spoken utterance from the audio. 

Results suggest that generally, SLMs slightly improve their results the more information is provided to the model. An exception to this is SALMONN which shows better results when it is only provided stress as opposed to stress and transcription. 

Our proposed model, \sftmodel, that results with $86.2$ and $87.6$ on \stressds and \stresspresso using the audio signal alone, on average (86.6) preserves its capabilities on \ssr in both scenarios on both datasets. It also demonstrates improvement in the oracle scenario when both transcription and stress are provided, perhaps due to the slight decrease in transcription abilities shown in~\Cref{tab:librispeech_wer}. Additionally, gpt-4o-audio, compared to its text-only counterpart in~\Cref{tab:reasoning_task} that yields $86.2$ and $83.6$ on \stressds and \stresspresso respectively, shows comparable performance when stress information is provided.

Notably, Gemini outperforms all other models by a wide margin when given ground-truth stress information. This, in contrast to its audio only performance that achieves less than $80.0$ \ssr. We attribute this gap to the poor precision in detecting stress as shown in~\Cref{tab:stress_detection}. 

Overall, these findings underscore the importance of precision and overall \ssd performance for improving \ssr.

\subsection{Open-ended sentence stress reasoning}
\label{appendix::open_ssr}
In~\Cref{tab:open_ssr} we provide additional results on the \stressds and \stresspresso datasets, evaluating the ability of leading SLMs to reason over stress meaning in an open-ended question format. Responses were evaluated by an LM-judge on a scale of $1$ to $5$, where a score of $5$ denotes a perfect capture of the intended meaning and a score of $1$ indicates a completely unrelated response. Notably, scores in the $2$ to $3$ range represent 'Poor' or 'Partial' comprehension. At these levels, models often produce vague or incomplete interpretations that miss the specific nuances of stress-driven ambiguity.
Results demonstrate a fundamental weakness in most current SLMs, which frequently yield scores in this lower range. Among the baselines, only Gemini-2.5-Pro shows more advanced reasoning, often reaching higher-tier scores. However, \sftmodel remains remarkably robust, demonstrating the ability to generalize across the stress meanings found in both datasets.

\begin{table}[h!]
\centering
\resizebox{\columnwidth}{!}{
\begin{tabular}{lcc}
\toprule
\multirow{2}{*}{\textbf{Model}} 
& \multicolumn{2}{c}{\textbf{Open SSR} $\uparrow$} \\
\cmidrule(lr){2-3}
& \textbf{\stressds} & \textbf{\stresspresso} \\
\midrule
Gemini-2.5-Pro & \textbf{3.82} & \underline{3.29} \\
gpt-4o-audio & 3.09 & 2.84 \\
Audio-Flamingo-3 & 2.62 & 2.39 \\
Qwen3-Omni-30B-Instruct & 2.51 & 2.42 \\
Qwen2Audio-7B-Instruct & 2.84 & 2.79 \\
SALMONN & 2.07 & 2.19 \\
LLaMA-Omni & 1.89 & 1.92 \\
Phi-4-multimodal-instruct & 2.23 & 2.73 \\
\midrule
\sftmodel (ours) & \underline{3.70} & \textbf{3.83} \\
\bottomrule
\end{tabular}
}
\caption{Open-ended stress reasoning (Open \ssr) performance on the \stressds and \stresspresso datasets.}
\label{tab:open_ssr}
\end{table}

\subsection{Ablation study}
\label{appendix:ablation}

We complement the ablation study results presented in~\Cref{tab:verifier_effectiveness} and~\Cref{tab:ablation} with results on \stresspresso benchmark. Results in~\Cref{tab:verifier_effectiveness_stresspresso} and~\Cref{tab:ablation_stresspresso} show similar trends to the ablation study in~\Cref{sec:ablation} supporting our design choices.

\begin{table}[t!]
    \centering
    \resizebox{\columnwidth}{!}{
    \begin{tabular}{cccccc}
    \toprule
    \multirow{2}{*}{\textbf{Verifier}} & \multirow{2}{*}{\textbf{\# Samples}} & \multicolumn{3}{c}{\textbf{SSD $\uparrow$}} & \textbf{SSR} $\uparrow$ \\
    \cmidrule(r){3-5} \cmidrule(r){6-6}
      &  & \textit{P.} & \textit{R.} & \textit{F1} & \textit{Acc.} \\
    \midrule
    \cmark & $\sim$4K & \textbf{81.1} & \underline{83.4} & \underline{82.3} & \underline{85.5}\\
    \xmark & $\sim$17K & 71.6 & 79.7 & 75.4 & 82.1 \\
    \xmark $\rightarrow$ \cmark & 17K$\rightarrow$4K & \underline{78.8} & \textbf{86.3} & \textbf{82.4} & \textbf{85.6}\\
    \bottomrule
    \end{tabular}}
    \caption{Effect of using \whistress verifier on model performance on \stresspresso.}
    \label{tab:verifier_effectiveness_stresspresso}  
\end{table}

\begin{table}[t!]
    \centering
    \resizebox{\columnwidth}{!}{
    \begin{tabular}{cccc|cccc}
    \toprule
    \textit{Train} & \textit{Stress} & \textit{Cas.} & \textit{Elab.} & \multicolumn{3}{c}{\textbf{SSD} $\uparrow$}  & \textbf{SSR} $\uparrow$ \\
    \cmidrule(r){5-7} \cmidrule(r){8-8}
    \textit{Enc.} & \textit{Det.} & \textit{Reas.} & \textit{Exp.} & \textit{P.} & \textit{R.} & \textit{F1} & \textit{Acc.}  \\
    \midrule
    \xmark & \cmark & \cmark & \cmark & 78.8 & 86.3 & 82.4 & 85.6 \\
    \cmidrule(r){1-8}
    \cmark & \xmark & \cmark & \cmark & 31.5 & \textbf{89.6} & 46.6 & 87.6 \\
    \cmark & \cmark & \xmark & \cmark & 71.7 & 82.5 & 76.7 & \textbf{88.6} \\
    \cmark & \cmark & \cmark & \xmark & \textbf{80.6} & \underline{88.2} & \textbf{84.2} & \underline{88.1} \\
    \cmidrule(r){1-8}
    \cmark & \cmark & \cmark & \cmark & \underline{79.9} & 87.2 & \underline{83.3} & \textbf{88.6} \\
    \bottomrule
    \end{tabular}}
    \caption{Ablating different training choices of \sftmodel on stress modeling capabilities, using staged-training on \stresspresso. We report Precision (P), Recall (R), F1, and Accuracy (Acc).}
    \label{tab:ablation_stresspresso}
\end{table}

\section{Prompts}
\label{appendix:prompts}
\subsection{Sentence stress reasoning evaluation}
\paragraph{Sentence stress reasoning accuracy.} The following prompt is used to query speech-aware LMs to evaluate the performance on the proposed end-to-end binary choice QA sentence stress reasoning.

\begin{styleprompt}[title=\ssr accuracy Prompt]
\ttfamily 
[audio]

Out of the following answers, according to the speaker's stressed words, what is most likely the underlying intention of the speaker?

1. [answer 1]

2. [answer 2]

Answer:
\end{styleprompt}

\paragraph{Open-ended sentence stress reasoning.} The following prompt is used to query speech-aware LMs to evaluate the performance on the open ended sentence stress reasoning task.

\begin{styleprompt}[title=Open-\ssr Prompt]
\ttfamily 
[audio]

According to the speaker's stressed words, what is most likely the underlying intention of the speaker?

Answer:
\end{styleprompt}

\paragraph{SSR Accuracy LM judge.} For the LLM-as-judge model we used the following prompt.
\begin{styleprompt}[title=LLM-as-judge \ssr Accuracy Prompt]
\ttfamily 
<system prompt>

You are a Speech-LM evaluator that helps evaluating models that have trouble in outputting a correct schema for an answer.
You are very good at outputting the correct schema according to the instructions.

INSTRUCTIONS:
Given a prompt with a question and possible answers that the Speech-LM received, and the output the model emmited, you are required to output the Speech-LM answer in a fixed format.

* The output should be aligned with the Speech-LM output, and should not include any additional information or context.

* The output should be a JSON object with a single key "answer" and a value that is the number of the correct answer according to the output of the Speech-LM.

* The answer should be an integer, either 1 or 2.
  
EXAMPLE 1:

INPUT TO Speech-LM:
    According to the intonation of the speaker, what is more probable?

    1. Yesterday, someone did not inform the speaker about the meeting.
    
    2. Someone did not inform the speaker about the meeting that occurred yesterday.

OUTPUT FROM Speech-LM:
    Someone did not inform the speaker about the meeting that occurred yesterday. Therefore, option 2 is more probable than option 1.

YOUR EXPECTED JSON OUTPUT:

\{"answer": 2\}

EXAMPLE 2:

INPUT TO Speech-LM:
    Question: According to the intonation of the speaker, what is more probable?
    Possible answers:

    1. Yesterday, someone did not inform the speaker about the meeting.
    
    2. Someone did not inform the speaker about the meeting that occurred yesterday.

OUTPUT FROM Speech-LM:
    Answer: 1. Yesterday, someone did not inform the speaker about the meeting.

YOUR EXPECTED JSON OUTPUT:
\{"answer": 1\}

<user prompt>

INPUT TO Speech-LM:

[input prompt]

OUTPUT FROM Speech-LM:

[speech lm output]

YOUR EXPECTED JSON OUTPUT:

\end{styleprompt}

\paragraph{Open-Ended SSR LM judge.} For the open question \ssr evaluation using LLM-as-judge, we used the following prompt.
\begin{styleprompt}[title=LLM-as-judge Open-\ssr Prompt]
\ttfamily 
<system prompt>

You are a Speech-LM evaluator. Your role is to evaluate open-ended answers produced by a Speech-LM.

INSTRUCTIONS:
You are given:
(1) The question the Speech-LM received.

(2) Reference information about the utterance:

    - Ground truth transcription.
    
    - Ground truth stressed words.
    
    - Ground truth intended meaning (based on the stressed words).

(3) The output the Speech-LM emitted. Your task is to output a score from 1 to 5 reflecting how well the Speech-LM answered the question,
  taking into account the reference information.

FOCUS OF EVALUATION:

- Does the Speech-LM's answer correctly capture the *intended meaning* of the speaker?

- Is the answer consistent with the stressed words and their pragmatic role?

- Is the answer relevant, coherent, and on-topic with respect to the question?

SCORING CRITERIA (1–5):
5 — Perfect:
    - Fully captures the ground truth intended meaning.

    - Consistent with the stressed words and their role in interpretation.

    - Clear, coherent, and directly answers the question.

4 — Strong:

    - Mostly captures the intended meaning, with only minor omissions or imprecision.

    - Generally consistent with stressed words.

    - Clearly relevant and coherent.

3 — Partial:

    - Contains some aspects of the intended meaning but is incomplete, vague, or partially off.

    - May miss some implications of stress or misinterpret nuances.

2 — Poor:

    - Mostly incorrect or missing the intended meaning.

    - Weak or incorrect use of information suggested by stressed words.

    - Only slightly related to the true intention.

1 — Very poor:

    - Completely incorrect, nonsensical, or unrelated to the question and reference meaning.

OUTPUT FORMAT (STRICT):

    * Output must be a JSON object.

    * The JSON object must contain a single key "answer".

    * "answer" must be an integer from 1 to 5.

    * Do not include explanations, reasoning, or any extra text.

    * Do not restate the answer, the question, or the reference; only output the JSON.

EXAMPLE (score = 5):

    QUESTION TO Speech-LM:
    
        According to the speaker's stressed words, what is most likely the underlying intention of the speaker?

REFERENCE INFORMATION:
        Ground truth transcription:
          "I asked you to call her yesterday."
          
        Ground truth stressed words:
        
          ["call"]
          
        Ground truth intended meaning:
        
          The speaker is emphasizing that the requested action was to call as opposed to any other action.

    OUTPUT FROM Speech-LM:

    The speaker is annoyed because the listener might have texted instead of calling, even though calling was specifically requested.

    YOUR EXPECTED JSON OUTPUT:

    \{"answer": 5\}

  EXAMPLE (score = 4):
  
    QUESTION TO Speech-LM:
    
        According to the speaker's stressed words, what is most likely the underlying intention of the speaker?

    REFERENCE INFORMATION:
    
        Ground truth transcription:
        
          "I asked you to call her yesterday."

        Ground truth stressed words:
        
          ["call"]

        Ground truth intended meaning:

          The speaker is emphasizing that the requested action was to call as opposed to any other action.

    OUTPUT FROM Speech-LM:
    
        The speaker is emphasizing that the listener needed to call her, but it doesn’t mention any contrast with other possible actions.

    YOUR EXPECTED JSON OUTPUT:
    
        \{"answer": 4\}

  EXAMPLE (score = 3):

    QUESTION TO Speech-LM:
    
        According to the speaker's stressed words, what is most likely the underlying intention of the speaker?

    REFERENCE INFORMATION:

        Ground truth transcription:
          "I asked you to call her yesterday."

        Ground truth stressed words:
        
          ["call"]

        Ground truth intended meaning:

          The speaker is emphasizing that the requested action was to call as opposed to any other action.

    OUTPUT FROM Speech-LM:

        The speaker is reminding the listener that they were supposed to contact her, though it's not clear how.

    YOUR EXPECTED JSON OUTPUT:
    
        \{"answer": 3\}

  EXAMPLE (score = 2):

    QUESTION TO Speech-LM:

        According to the speaker's stressed words, what is most likely the underlying intention of the speaker?

    REFERENCE INFORMATION:
    
        Ground truth transcription:
        
          "I asked you to call her yesterday."
        Ground truth stressed words:
        
          ["call"]
          
        Ground truth intended meaning:
        
          The speaker is emphasizing that the requested action was to call as opposed to any other action.

    OUTPUT FROM Speech-LM:
    
        The speaker seems to be reminding the listener that they were supposed to reach out to her, but it's unclear what method of contact was important.

    YOUR EXPECTED JSON OUTPUT:
    
        \{"answer": 2\}

  EXAMPLE (score = 1):
  
    QUESTION TO Speech-LM:
    
        According to the speaker's stressed words, what is most likely the underlying intention of the speaker?

    REFERENCE INFORMATION:
    
        Ground truth transcription:
          "I asked you to call her yesterday."
          
        Ground truth stressed words:
        
          ["call"]
          
        Ground truth intended meaning:
        
          The speaker is emphasizing that the requested action was to call as opposed to any other action.

    OUTPUT FROM Speech-LM:
    
        The speaker is stressing "yesterday" meaning the main intention is to point out that the listener acted too late.

    YOUR EXPECTED JSON OUTPUT:
        \{"answer": 1\}

<user prompt>

  QUESTION TO Speech-LM:
  
  [input prompt]
  
  REFERENCE INFORMATION:
  
    Ground truth transcription:
    
      [gt transcription]
      
    Ground truth stressed words:
    
      [gt stressed words]
      
    Ground truth intended meaning:
    
      [gt intended meaning]
      
  OUTPUT FROM Speech-LM:
  
  [audio llm output]
  
  YOUR EXPECTED JSON OUTPUT:
  
\end{styleprompt}

\subsection{Sentence stress detection evaluation}
\paragraph{Stress detection task.} We use the following prompt to assess the $SSD$ task.
\begin{styleprompt}[title=Speech-Aware LM \ssd Prompt]
\ttfamily 
[audio]

The speaker said "[transcription]". 

According to the audio, what words did the speaker stress?

Answer format: [stressed\_word\_1, ...]

Answer: 
\end{styleprompt}

\paragraph{Detection LM judge.} For the LLM-as-judge model we used the following prompt.
\begin{styleprompt}[title=LLM-as-judge \ssd Prompt]
\ttfamily 
<system prompt>
  You are a Speech-LM evaluator that helps evaluating models that have trouble in outputting a correct schema for an answer.
  You are very good at outputting the correct schema according to the instructions.
  
  INSTRUCTIONS:
  Given a prompt with a question that the Speech-LM received, and the output the model emitted, you are required to output the Speech-LM answer in a fixed format.
  
  * The output should be aligned with the Speech-LM output, and should not include any additional information or context.
  
  * The output should be a JSON object with a single key "answer" and a value that is a list of words according to the output of the Speech-LM.
  
  * The answer should be a list of strings.
  
  * If the model mistakenly outputs two or more words as a single word, you should split them into separate words.
  
  EXAMPLE:
  
    INPUT TO Speech-LM:
    
        The speaker said "What a lovely day we have". 
        
        According to the audio, what words did the speaker stress?
        
        Answer format: [stressed\_word\_1, ...]
        
        Answer: 
        
    OUTPUT FROM Speech-LM:
    
        The speaker stressed: ["lovely", "we have"].
        
    YOUR EXPECTED JSON OUTPUT:
    
    \{"answer": ["lovely", "we", "have"]\}

<user prompt>

INPUT TO Speech-LM:

[input prompt]

OUTPUT FROM Speech-LM:

[speech lm output]

YOUR EXPECTED JSON OUTPUT:

\end{styleprompt}

\subsection{Reasoning analysis prompts}
\label{subsec:reasoning_analysis_prompts}

\paragraph{Transcription and stress as input.} The prompt evaluates whether additional context helps sentence stress understanding. In case the evaluated model is an LM, the audio placeholder is not used.

\begin{styleprompt}[title=Sentence Stress Reasoning - stress and transcription input]
\ttfamily 
[Audio]

Question: 

Given that a speaker said: "[transcription]", and stressed the words: [stressed words]. 

Out of the following answers, what is most likely the underlying intention of the speaker?

Possible answers:

1. [answer 1]

2. [answer 2]

Answer: 
\end{styleprompt}

\paragraph{Stress as input.} The prompt evaluates whether only the stressed words helps sentence stress understanding, since ASR is a fundamental task for speech-aware LMs.

\begin{styleprompt}[title=Sentence Stress Reasoning - stress input]
\ttfamily 
[Audio]

Question:

Out of the following answers, given that the speaker stressed the words: [stressed words]. 

What is most likely the underlying intention of the speaker?

Possible answers:

1. [answer 1]

2. [answer 2]

Answer:
\end{styleprompt}

\subsection{Training prompts}
\label{appendix:training_prompts}
\paragraph{End-to-end task.} The following prompt guide the model to precisely choose the speaker's intended meaning based on stressed words.

\begin{styleprompt}[title=End to end reasoning]
\ttfamily
[Audio]

Out of the following answers, according to the speaker's stressed words, what is most likely the underlying intention of the speaker?

1. [answer 1]
    
2. [answer 2]

Answer:
\end{styleprompt}

\begin{styleprompt}[title=Expected Answer Format]
\ttfamily
[answer label]. [correct answer]
\end{styleprompt}

\paragraph{Elaborated answer task.} The model is required to first explain its reasoning and then answer.

\begin{styleprompt}[title=Elaborated Answer Prompt]
\ttfamily
[Audio]

According to the speaker's stressed words, what is the speaker's underlying intention? 

1. [answer 1]
    
2. [answer 2]

Elaborate, then answer in the following way: "answer\_number. correct\_answer"
\end{styleprompt}

\begin{styleprompt}[title=Expected Answer Format]
\ttfamily
[description]. Therefore, the correct answer is: [answer label]. [correct answer]
\end{styleprompt}
\paragraph{Cascade reasoning task.} This prompt encourages the model to reason based on the stressed words and transcription before answering.

\begin{styleprompt}[title=Stress Detection Reasoning Prompt]
\ttfamily
[Audio]

The speaker stressed some words. What is the speaker trying to communicate? 

1. [answer 1]
    
2. [answer 2]

Think about the transcription and the stressed words. Then, answer like this: "answer\_number. correct\_answer"
\end{styleprompt}

\begin{styleprompt}[title=Expected Answer Format (Format 7)]
\ttfamily
The speaker said "[transcription]" and emphasized "[stressed words]".

Therefore, the correct answer is: [answer label]. [correct answer]
\end{styleprompt}

\paragraph{Stress detection task.} This prompt focuses only on detecting which words were emphasized.

\begin{styleprompt}[title=Stress detection prompt]
\ttfamily
[Audio]

The speaker said "[transcription]". 

According to the audio, what words did the speaker stress?

Answer format: [stressed\_word\_1, ...]

Answer: 
\end{styleprompt}

\begin{styleprompt}[title=Expected Answer Format]
\ttfamily
[stressed words]
\end{styleprompt}


\subsection{Data generation pipeline}
\label{appendix:data_generation_textual_samples}
\paragraph{Text sample generation.} The agentic process used to create the textual examples will be open-sourced with the full yml configurations and code.

\paragraph{Metadata.} We use 10 sentence types: statement, question, command, exclamation, request, suggestion, invitation, offer, opinion and warning. Additionally, a domain and a topic are injected into the agent generation prompt out of a list of $32$ domains and $110$ topics corresponding to the domains, generated by gpt-4o.
\begin{styleprompt}[title=Domains and topics]
\small
\ttfamily
Art \& Culture:

-The Renaissance Period

-Street Art and Graffiti

-The Influence of Jazz Music

-The Role of Film in Society

-Modern Architecture Trends

-Art as Political Protest

-The Influence of Surrealism

-Cultural Appropriation in Fashion

-Folk Music and Tradition

-The Globalization of Art Markets

-The Role of Museums in Preserving Culture

-The Influence of Classical Music on Modern Genres

-Digital Art and Its Rising Popularity

-The Evolution of Dance Across Cultures

-The Impact of the Internet on Art Consumption

Business:

-Startup Culture and Innovation

-Corporate Social Responsibility

-Leadership Styles and Management

-The Gig Economy

Economics:

-Income Inequality

-Global Trade Wars

-Cryptocurrency and Digital Economy

-The Future of Work and Automation

Education:

-Online Learning Platforms

-Special Education Needs

-The Role of Arts in Education

-STEM Education for Girls

Engineering:

-Renewable Energy Engineering

-Aerospace Innovations

-Civil Engineering and Urban Planning

-Robotics and Automation

Environment:

-Ocean Pollution and Marine Life

-Conservation of Endangered Species

-Urban Green Spaces

-Sustainable Agriculture Practices

Food \& Culinary Arts:

-The Science of Baking

-Global Cuisine Trends

-Farm-to-Table Movement

-Veganism and Plant-Based Diets

Health \& Medicine:

-Mental Health Awareness

-The Rise of Telemedicine

-Nutrition and Lifestyle Diseases

-Vaccine Development Processes

History:

-The Industrial Revolution

-Ancient Civilizations and Their Contributions

-World Wars and Their Consequences

-The History of Human Rights

Law:

-Intellectual Property Rights

-The Justice System and Reforms

-International Law and Human Rights

-Cyber Law and Internet Regulations

Literature:

-Dystopian Novels and Society

-The Impact of Shakespeare

-The Evolution of Poetry

-Modern Graphic Novels

Philosophy:

-Existentialism and Modern Thought

-The Ethics of Artificial Intelligence

-Eastern vs. Western Philosophical Traditions

-The Philosophy of Science

Politics:

-Globalization and Nationalism

-Election Systems and Voter Rights

-The Role of the United Nations

-Immigration Policies and Refugee Crisis

Psychology:

-Cognitive Behavioral Therapy

-The Psychology of Social Media

-Child Development Stages

-Emotional Intelligence in Leadership

-The Impact of Stress on Health

Religion:

-Comparative Religion Studies

-Secularism and Society

-Rituals and Traditions

-The Role of Religion in Politics

Science:

-CRISPR and Genetic Engineering

-Climate Change and Its Impact on Biodiversity

-Space Exploration and Mars Colonization

-Nanotechnology in Medicine

Sociology:

-Urbanization and Its Challenges

-The Dynamics of Family Structures

-Gender Roles in Society

-The Sociology of Religion

Sports:

-The Science of Sports Performance

-Gender Equality in Sports

Technology:

-Artificial Intelligence and Ethics

-Quantum Computing Advancements

Travel \& Tourism:

-Eco-Tourism and Sustainability

-Cultural Heritage Sites

Anthropology:

-Cultural Practices of Hunter-Gatherer Societies

-The Evolution of Human Speech and Language

Astronomy:

-The Birth and Death of Stars

-The Formation of Black Holes

Cryptography:

-The Evolution of Cryptographic Algorithms

-Quantum Cryptography and Secure Communication

Fashion:

-The Rise of Fast Fashion and Its Environmental Impact

-Fashion Icons Who Changed History

Gaming:

-The Psychology of Video Game Addiction

-The Rise of Esports and Competitive Gaming

Geopolitics:

-The Role of Natural Resources in International Conflict

-Geopolitical Impacts of Climate Change

Mathematics:

-Chaos Theory and its Applications

-The Role of Mathematics in Predictive Modeling

Performing Arts:

-The Evolution of Contemporary Dance

-The Role of Theater in Political Activism

Photography:

-The Evolution of Documentary Photography

-The Role of Photography in Social Movements

Space Exploration:

-The Privatization of Space Travel

-The Role of Satellites in Modern Communication

Visual Arts:

-The Impact of Technology on Visual Arts

-The Role of Photography in Modern Art

Urban Studies:

-The Role of Public Transportation in Urban Growth

-Urbanization and Its Impact on Housing
\end{styleprompt}

\section{Broader impact}
\label{apendix:broader_impact}
As with any advancement in language models, our work carries the potential for both positive and unintended consequences. By equipping models with the ability to interpret the meaning conveyed through stress patterns in speech, we aim to foster more efficient and nuanced communication systems. Such capabilities may particularly benefit populations who rely heavily on prosody for meaning - such as individuals with hearing disabilities, language learners, etc and ultimately contribute to more accessible and context-aware AI systems.

\section{Usage of LLMs}
\label{apendix:llm_usage}
As part of this research study we use LLMs for two main reasons: (i) Writing helper, e.g., improving writing style; (ii) LLM-as-a-Judge. For writing style we mainly use Chat-GPT, while for LLMs-as-a-Judge we use gpt4o.
\end{document}